\documentclass[conference]{IEEEtran}
\UseRawInputEncoding
\usepackage{cite}
\usepackage{amsmath,amssymb,amsfonts}
\usepackage{algorithmic}
\usepackage{graphicx}
\usepackage{textcomp}
\usepackage{xcolor}
\usepackage{float}
\usepackage{hyperref}
\usepackage{subfig}
\usepackage{multirow}
\usepackage{caption}

\def\BibTeX{{\rm B\kern-.05em{\sc i\kern-.025em b}\kern-.08em
    T\kern-.1667em\lower.7ex\hbox{E}\kern-.125emX}}
    
\begin{document}
\title{Analyzing Multispectral Satellite Imagery of South American Wildfires Using Deep Learning}
\author{\IEEEauthorblockN{Christopher Sun}
\IEEEauthorblockA{\textit{Monta Vista High School} \\
Cupertino, CA, United States}}

\maketitle

\begin{abstract}
Since frequent severe droughts are lengthening the dry season in the Amazon Rainforest, it is important to detect wildfires promptly and forecast possible spread for effective suppression response. Current wildfire detection models are not versatile enough for the low-technology conditions of South American hot spots. This deep learning study first trains a Fully Convolutional Neural Network on Landsat 8 images of Ecuador and the Galapagos, using Green and Short-wave Infrared bands to predict pixel-level binary fire masks. This model achieves a 0.962 validation F2 score and a 0.932 F2 score on test data from Guyana and Suriname. Afterward, image segmentation is conducted on the Cirrus band using K-Means Clustering to simplify continuous pixel values into three discrete classes representing differing degrees of cirrus cloud contamination. Three additional Convolutional Neural Networks are trained to conduct a sensitivity analysis measuring the effect of simplified features on model accuracy and train time. The Experimental model trained on the segmented cirrus images provides a statistically significant decrease in train time compared to the Control model trained on raw cirrus images, without compromising binary accuracy. This proof of concept reveals that feature engineering can improve the performance of wildfire detection models by lowering computational expense. 
\end{abstract}


\section{Introduction}
In places like the Amazon Rainforest, where severe droughts are occurring more frequently and lengthening the dry season, interconnected ecosystems are threatened by more widespread occurrences of wildfires (Aragão et al. 2018) \cite{intro}. It is not only important to respond to wildfires promptly using well-informed allocation of rescue resources to minimize waste, but also to forecast active wildfires before they become inextinguishable. Researchers have made progress towards this task by applying remote sensing techniques and machine learning algorithms on to detect wildfires. Matson and Holben (1987) used Dozier's method to extract pixel-level information from NOAA data of regions in Brazil \cite{matson holben}. Schroeder et al. (2016) used the Near-Infrared and Short-Wave Infrared bands from the Landsat 8 satellite to classify fire-affected pixels \cite{schroeder}. Khryashchev and Larionov (2020) used RGB images and data augmentation for wildfire detection \cite{khryashchev}. Most recently, Pereira et al. (2021) introduced three deep convolutional neural network architectures for active wildfire detection, using these models to predict the overhead pixel-level labels of a 256 x 256 pixel patch of land \cite{pereira}. However, these automated research designs have not conducted feature engineering apart from selecting spectral bands, which, coupled with complex model architectures with tens of millions of parameters, results in a high computational cost.
\\\\
\indent This paper introduces a fully convolutional neural network (FCN) with skip connections that uses 3-channel inputs to predict corresponding fire masks. The paper then describes a clustering approach to image segmentation of cirrus images. The results of three additional convolutional models reveal that fires can be detected accurately using simplified features, which is useful when computational resources are scarce.
\begin{figure}[H]
    \centering
    \includegraphics[scale=0.1]{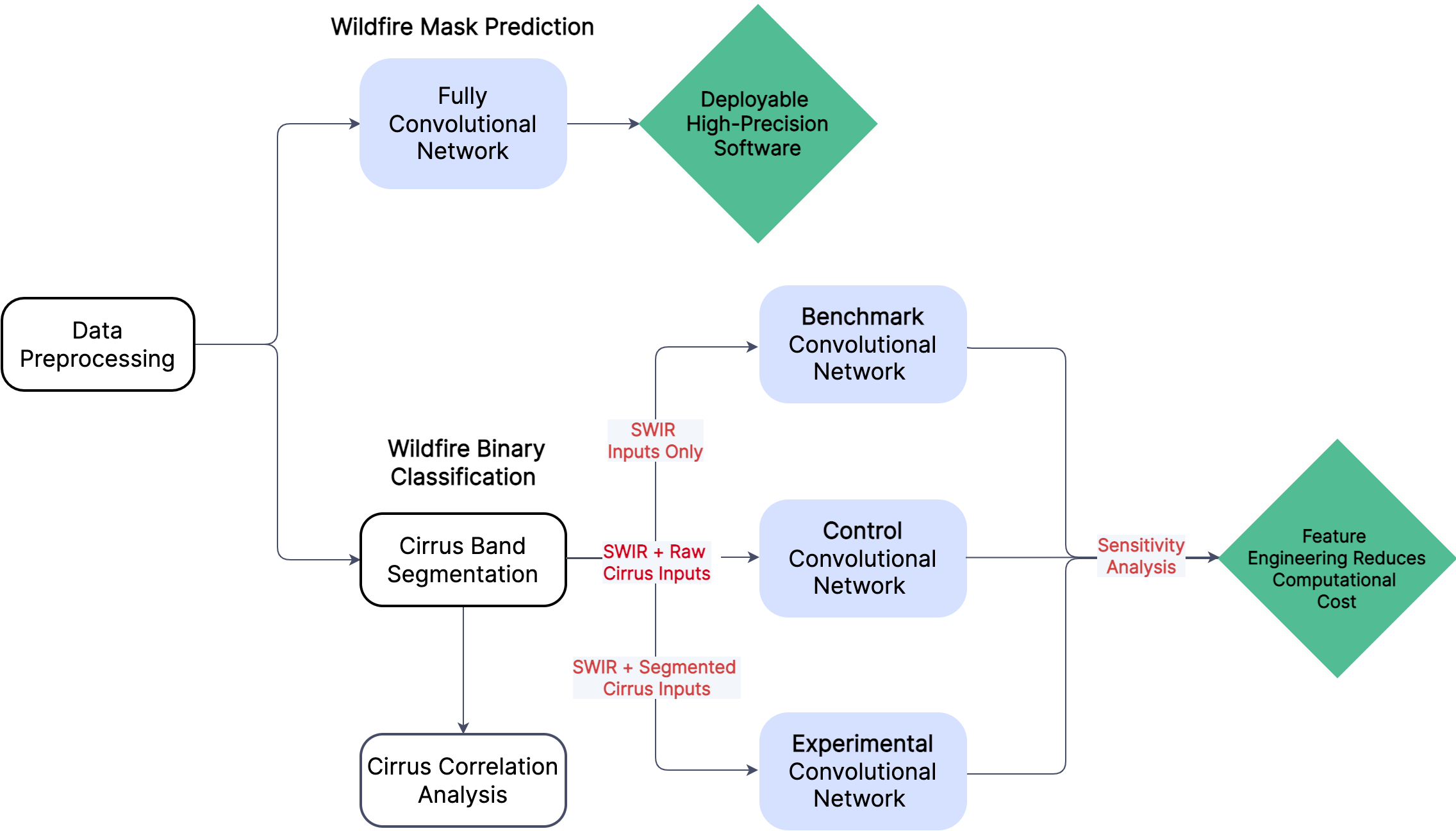}
    \caption{Experimental Design}
\end{figure}
\section{Materials and Data}
The data used for this research was a subset of a large publicly available database of Landsat 8 satellite images \cite{data set} processed by Pereira et al. (2021) \cite{pereira}. Aerial images of Ecuador and the Galapagos were selected to train the model, and images of Guyana and Suriname were used as test data for a final model evaluation. The data consisted of 14,274 images containing at least one pixel with an active fire, and 10,685 images containing no fires. The former set of images will be referred to as the ``fire'' images and the latter as the ``non-fire'' images. Each image had dimensions 128$\times$128$\times$10 and was paired with its corresponding binary mask that indicated the presence of absence of fire in the region denoted by each pixel. These 128$\times$128$\times$1 masks were either manually annotated or generated according to the conditions described in Schroeder et al. (2016) \cite{schroeder}. 
\begin{figure}[H]
    \centering
    \includegraphics[scale=0.3]{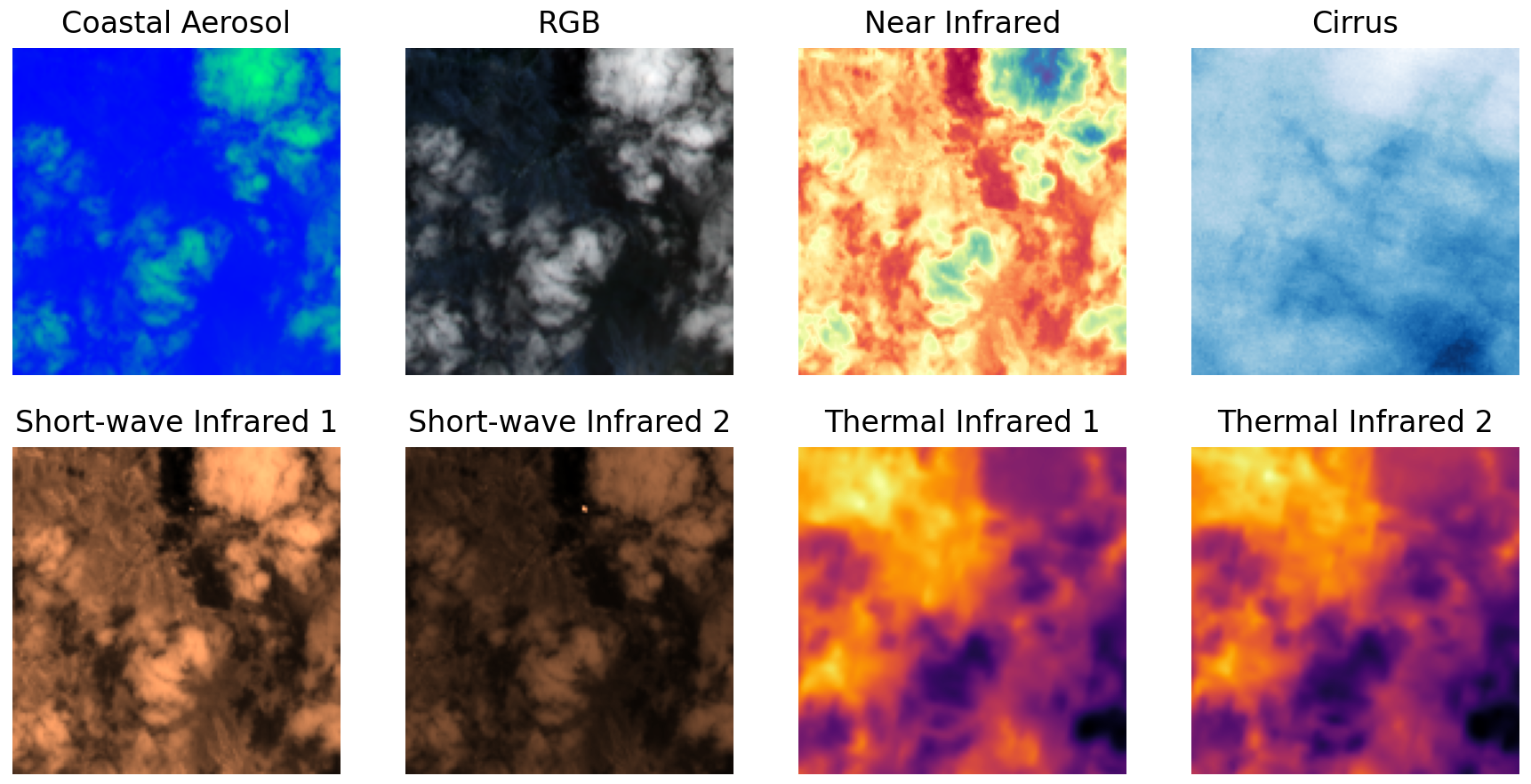}
    \caption{The data set included images with 10 spectral bands. Each band is shown, with RGB visualized as one band.}
\end{figure}
The models in this paper were trained using the NVIDIA Tesla P100 GPU. 
\section{Methods}
\subsection{Fully Convolutional Neural Network}
\subsubsection{Neural Network Design}
The ultimate modeling task was to extract features from each image to predict the corresponding mask. Accomplishing this would not require all ten bands of data, as demonstrated by Pereira et al. (2021), who used only three bands and generated good results \cite{pereira}. Consequently, only the Green and Short-wave Infrared bands were selected as inputs. The deep learning model assembled for this task was based on the U-Net\footnote{The Mask R-CNN was not used because most masks contained very few fire pixels, so the primary task was not instance segmentation.} \cite{unet}. Instead of the four downsampling and upsampling convolutional blocks in the U-Net, a shallower model consisting of only three downsampling and upsampling convolutional blocks was built, with a smaller number of filters per block as well. As a result, the model contained only 2,138,785 trainable parameters, which is considerably less than the 34,525,121-parameter model designed by Pereira et al. (2021) \cite{pereira}. During the downsampling phase, 2D convolution layers enlarge the number of channels, while maximum pooling reduces dimensionality along the first and second axes. During the upsampling phase, deconvolution layers reduce the number of channels and increase dimensionality along the first and second axes. The image is encoded, passed through a bottleneck phase, and decoded by the model. Skip connections between the convolutional blocks provide feature information to later layers that might be lost from downsampling (Drozdzal et al. 2016) \cite{skip connections}. A final 1$\times$1 convolution layer generates the 128$\times$128$\times$1 output mask. 
\begin{figure}[H]
    \centering
    \includegraphics[scale=0.2]{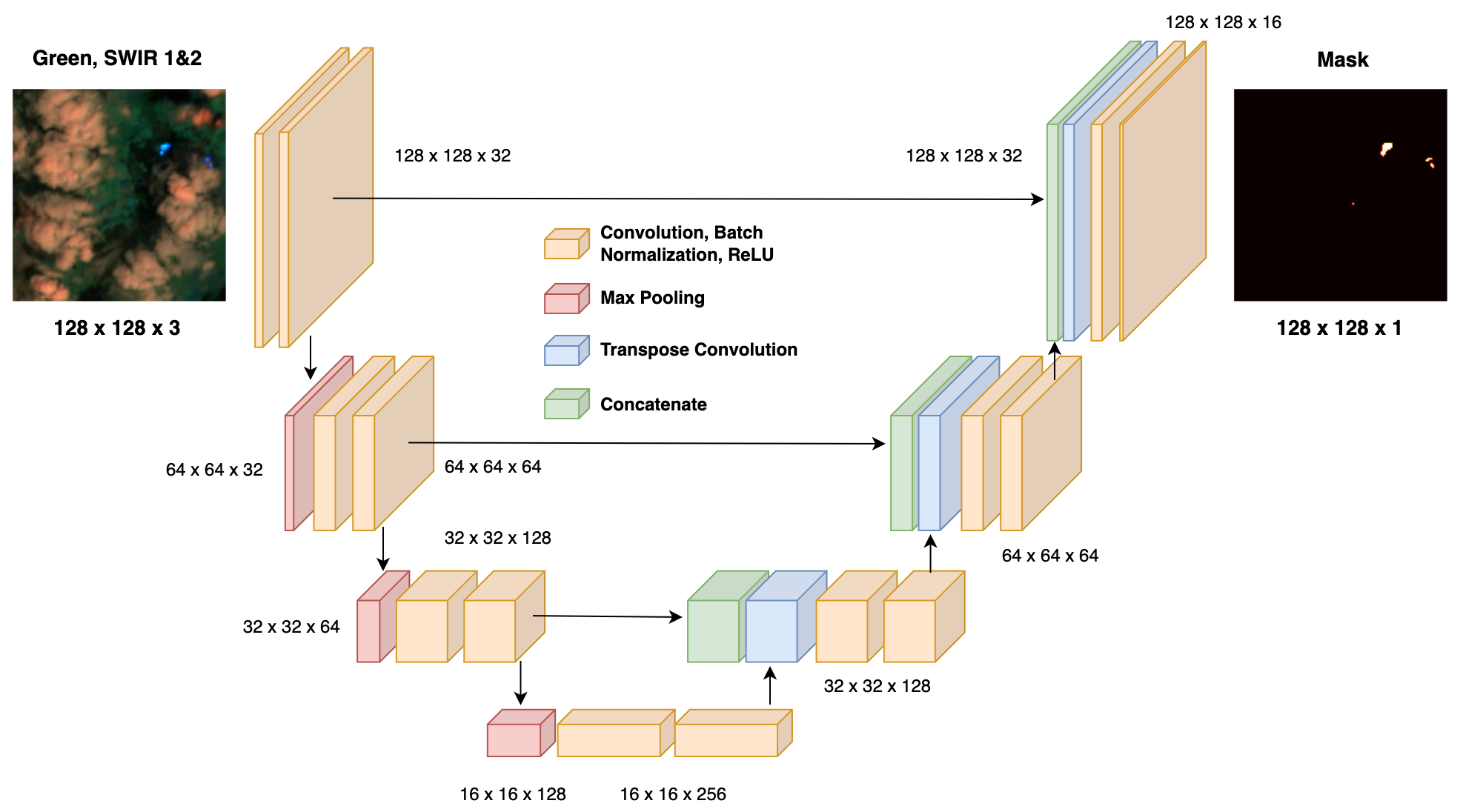}
    \caption{The architecture of the Fully Convolutional Model used for mask prediction.}
\end{figure}
\subsubsection{Loss and Metric Functions} \label{metrics}
Due to the class imbalance in the image masks, the model required a weighted loss function. The standard binary cross-entropy loss function used for binary classification was adjusted by a factor of $\mathbf{p_c}$, called the class weight. This class weight was computed using: \[\mathbf{p_c} = \frac{\textnormal{Total\;number\;of\;pixels\;that\;equal\;0}}{\textnormal{Total\;number\;of\;pixels\;that\;equal\;1}}\] Accordingly, the loss function was defined as: \small\[L(\theta) = \frac{1}{m}\sum^{m}_{i=1}\sum^{k}_{n=1}\left[\mathbf{p_c}\cdot y^{n}\log{(\hat{y}^n)}+(1-y^n)\log{(1-\hat{y}^n)}\right]^{i},\] 
\normalsize	where $k$ is the number of pixels in each flattened mask, $y^{n}$ is the ground truth classification of the $n^{th}$ pixel, $\hat{y}^n$ is the prediction value of the $n^{th}$ pixel, and a summation is applied over every pixel of the $i^{th}$ image for a total of $m$ images in each mini-batch. 
\\\\
The F-beta score with a beta value of 2 was used as a performance metric for the FCN\footnote{As the $F_1$ score is more widely used, the $F_1$ score will also be provided in Section \ref{model results} to allow for comparison with other published models.}. The $F_2$ is a weighted harmonic mean of the precision and recall, with more weight given to the recall. Maximizing recall was prioritized over maximizing precision, since false negatives are more detrimental than false positives during wildfire detection (Arana-Pulido et al. 2018) \cite{fn}. Let precision $P = \frac{tp}{tp + fp}$ and let recall $R = \frac{tp}{tp + fn}$. \[F_2 = \frac{5PR}{4P+R}. = \frac{tp}{tp+0.2fp+0.8fn}.\] 
\subsubsection{Training} \label{training}
The FCN was optimized using Adam optimization (Kingma and Ba 2014) \cite{adam} with an exponentially decaying learning rate. The learning rate $\alpha$ as a function of the epoch of training, $t$, was \[\alpha(t) = \alpha_0\,\textbf{e}^{-0.1t},\] where $\alpha_0$ was 1E-3 and 0.1 was a hyperparameter. The ReLU activation function was used throughout the model, with the exception of the last layer which used the sigmoid function. Batch Normalization was employed on the channels axis as a regularization technique \cite{batchnorm}. The model was trained for 100 epochs with a mini-batch size of 32 images. The 14,274 fire images were used to train the neural network on an 85-15\% train-validation split.
\subsection{Cirrus Band Analysis}
Environmental research has shown that subtleties in atmospheric responses to wildfires provide information about the sources of the fires themselves. Veselovskii et al. (2021) highlighted the association between wildfire smoke and cirrus clouds, using LiDAR observations to find that that smoke properties such as surface area, volume, and concentration could be extracted simultaneously with cirrus properties \cite{cirrus background}. 
\\\\
\indent This idea motivated another avenue of experimentation: extracting features from the cirrus bands as part of sensitivity analysis determining the effect of feature engineering on computational expense and accuracy. 
\subsubsection{Cirrus Data Preparation}
Since not all images in the data set contained the cirrus cloud contamination desirable for the analysis, images that contained cirrus clouds first needed to be selected. The criteria for determining whether an image contained cirrus clouds were experimentally decided. Visual observation of the cirrus bands confirmed that images with cirrus clouds had a much larger pixel value range than images without cirrus clouds. Accordingly, the minimum pixel value threshold for an image to contain cirrus clouds was set as 500.\footnote{In most cirrus images, raw pixel values hovered around 5,000.} 5,420 of the 24,959 original images satisfied this condition, with 2,152 fire images and 3,268 non-fire images.
\subsubsection{K-Means Clustering for Cirrus Image Segmentation}
Image segmentation was conducted to simplify the features of the cirrus clouds. For each image, a $128^{2}\times$3 array consisting of each pixel value along with the pixel's coordinates on the image, was fed into the K-Means Clustering algorithm with a K-value of 3.
\subsubsection{Clustering Hypothesis} It was hypothesized that if a K-value of 3 was used, the clustering algorithm would identify spatial regions corresponding to degrees of cloud contamination: ``dense'' cirrus, ``scattered'' cirrus, and ``no'' cirrus. 
\subsection{Cirrus-to-Wildfire CNN} \label{cirrus model}
\subsubsection{Sensitivity Analysis}
To test the hypothesis that feature simplification could reduce the train time yet preserve the accuracy of the wildfire detection algorithm, three convolutional neural networks were constructed using the same architecture but trained on different data features. 
\begin{itemize}
    \item \textbf{Benchmark Model (Only SWIR Data)} The benchmark model was trained only on the two Short-wave Infrared bands.
    \item \textbf{Control Model (SWIR and Raw Cirrus Data)} The control model was trained on the two Short-wave Infrared bands stacked depthwise with the raw Cirrus band.
    \item \textbf{Experimental Model (SWIR and Segmented Cirrus Data)} The experimental model was trained on the two Short-wave Infrared bands stacked depthwise with the segmented Cirrus band generated using clustering.
\end{itemize}
\subsubsection{Model Architecture}
All three of the above models used the same architecture, hyperparameters, and random seed. The models contained 1,824,937 trainable parameters. Because most parameters were concentrated in the final two convolution layers and the first fully connected layer, L2 Regularization with a lamdba value of 3E-3 was used in the final two convolution layers, and dropout regularization with a dropout probability of 0.3 was used before flattening the outputs of the last convolution. The fully connected layers were also regularized with dropout, with dropout probabilities decreasing by layer from 0.3 to 0.1. The 5,420 aforementioned images were used to train the neural network on an 80-20\% train-validation split.
\begin{figure}[H]
    \centering
    \includegraphics[scale=0.2]{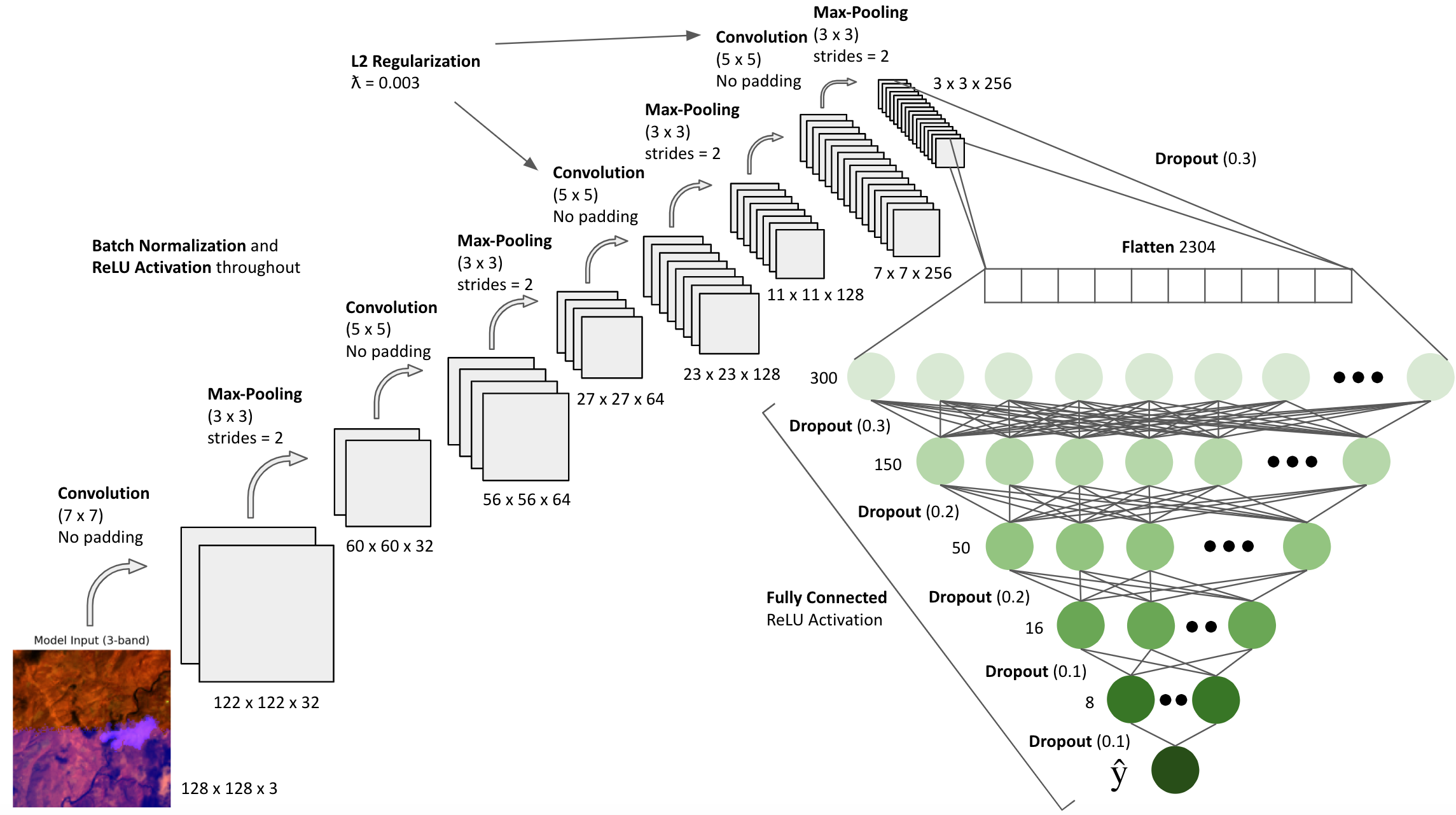}
    \caption{The architecture of the convolutional models used for the sensitivity analysis.}
\end{figure}

\section{Results}
\subsection{Fully Convolutional Neural Network}
\subsubsection{Learning Curves}
\begin{figure}[H]
    \centering
    \includegraphics[scale=0.3]{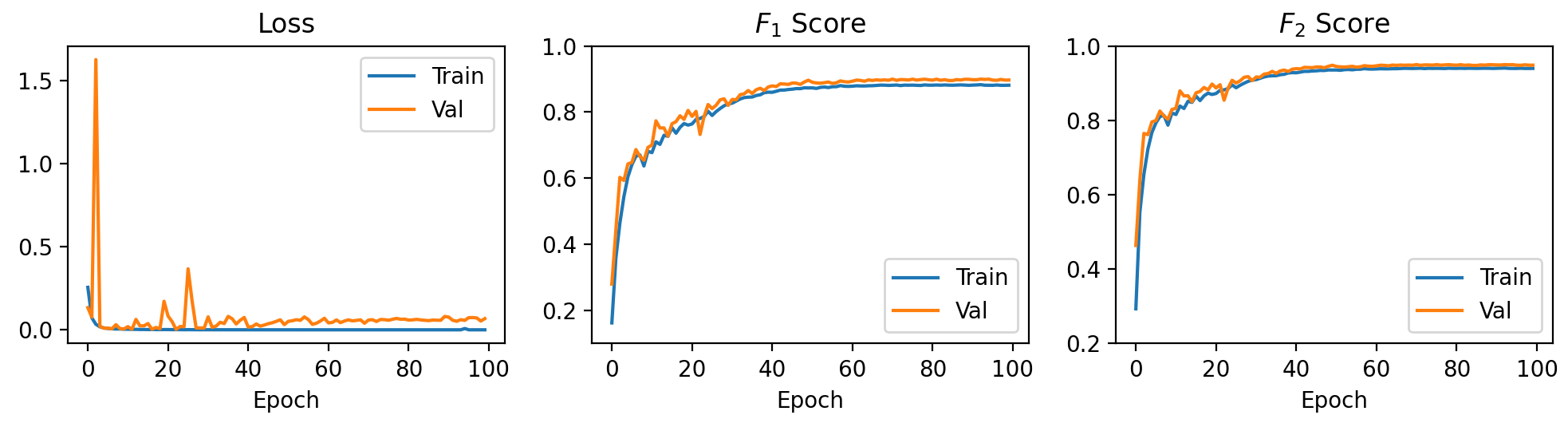}
    \caption{The performance of the FCN over 100 epochs.}
\end{figure}
The FCN converged quickly on the train set while the validation loss experienced some noise until around epoch 50. After epoch 50, the validation loss seemed to subtly increase, implying slight overfitting, but this was interestingly not reflected in the validation $F_1$ and $F_2$ Scores.
\subsubsection{Metrics and Confusion Matrix} \label{model results}
\begin{table}[H]
\caption{Confusion Matrix}
\begin{tabular}{l|l|l|l}
\cline{2-3} \\[-1em]
                                         & Actual: Fire & Actual: No Fire &                                 \\ \hline \\[-1em]
\multicolumn{1}{|l|}{Predicted: Fire}    & TP = \textbf{18,691}  & FP = \textbf{2,607}      & \multicolumn{1}{l|}{21,298}     \\ \hline \\[-1em]
\multicolumn{1}{|l|}{Predicted: No Fire} & FN = \textbf{277}     & TN = \textbf{35,072,953} & \multicolumn{1}{l|}{35,073,230} \\ \hline \\[-1em]
                                         & 18,968       & 35,075,560      & \multicolumn{1}{l|}{35,094,528} \\ \cline{2-4} 
\end{tabular}
\end{table}

The FCN achieved a 0.962 $F_2$ score and a 0.928 $F_1$ score on the validation data set, corresponding to a Precision of 0.878 and a Recall of 0.989. This means that the model very rarely failed to identity an existing fire, but was occasionally too sensitive, flagging non-fire pixels as fire pixels. 
\subsubsection{Performance Visualizations}
The following figure shows the predicted masks when there are a small number and large number of fire pixels in the mask, respectively. The third row of (b) shows that the model failed to identify nonfire pixels when they were surrounded by fire pixels. This over-labeling of fires was, however, not observed when the mask had only a fire pixels. 
\begin{figure}[H] 
    \centering
    \subfloat[Small number of fire pixels. Fire pixels are boxed.\label{1a}]{%
    \includegraphics[scale=0.22]{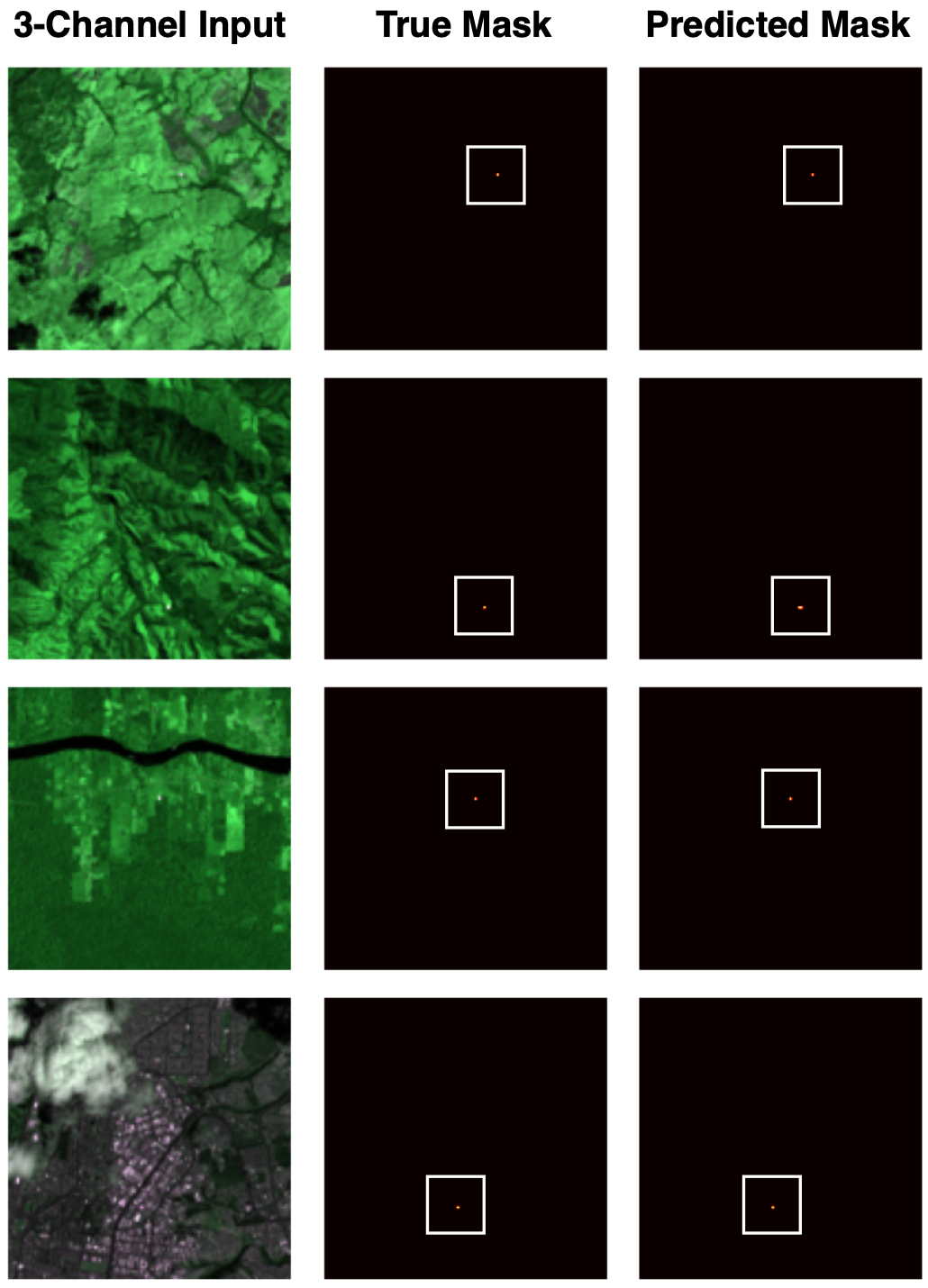}}
    \hfill
    \subfloat[Large number of fire pixels.\label{1b}]{%
    \includegraphics[scale=0.22]{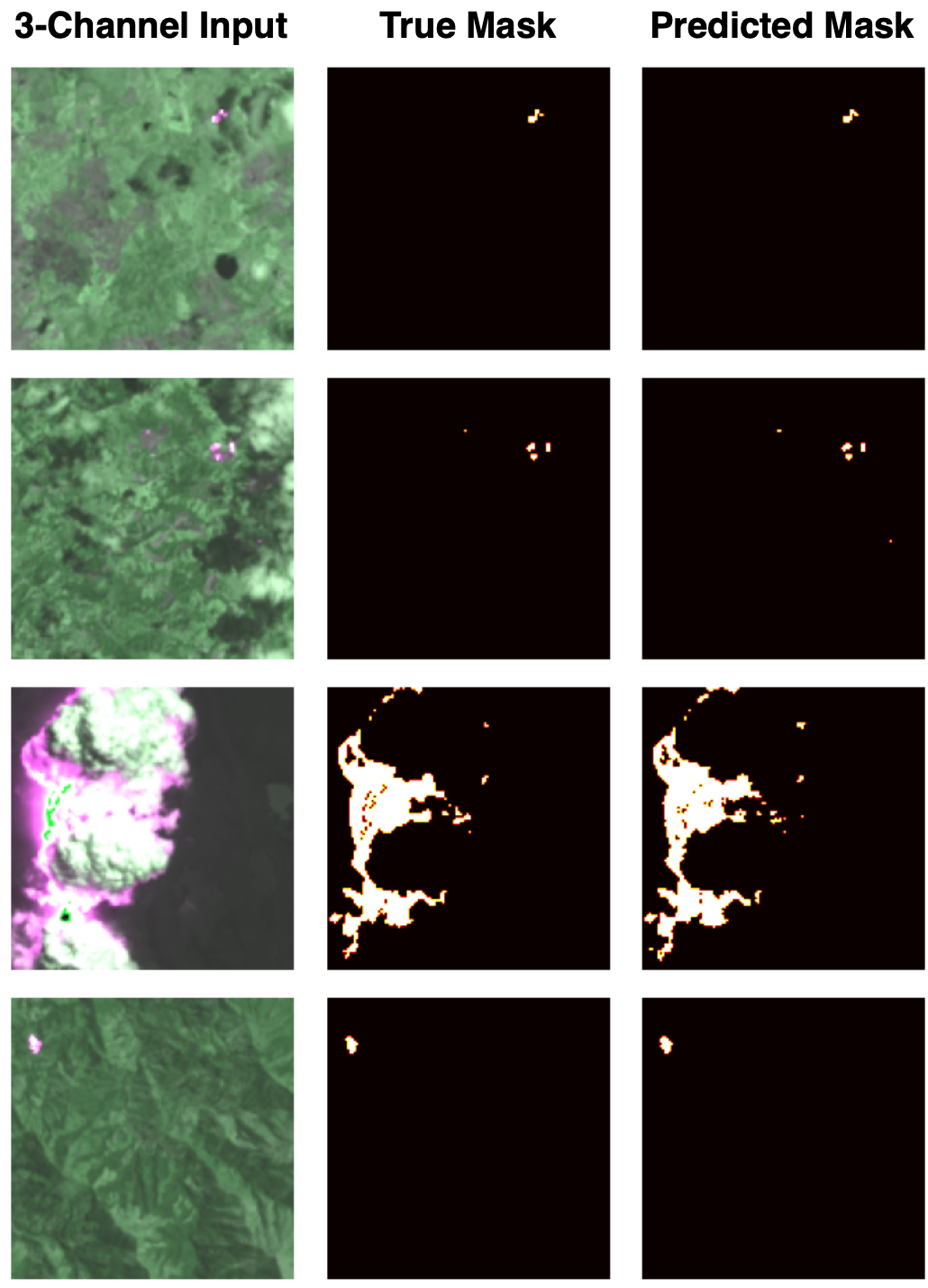}}
    \caption{(a), (b) The predicted masks resemble the true masks to a large degree. For images with more fire pixels, the model tends to predict more false positives.}
\end{figure}
\subsubsection{Model Testing Scenarios} \label{test data}
\paragraph{Geographical Test} The FCN was tested on 2,000 images from Guyana and Suriname and achieved a 0.934 $F_2$ score and a 0.875 $F_1$ score. These results confirm that the FCN is versatile, generalizing well to geographical regions that differ in topography and climate. 
\paragraph{Type II Error Test} The FCN was also tested on 2,000 non-fire images. Only 13 pixels were predicted to contain wildfires, implying an extremely low rate of false positives, at 3.967E-5\%.
\subsection{Cirrus Band Segmentation}
\begin{figure}[H]
    \centering
    \includegraphics[scale=0.3]{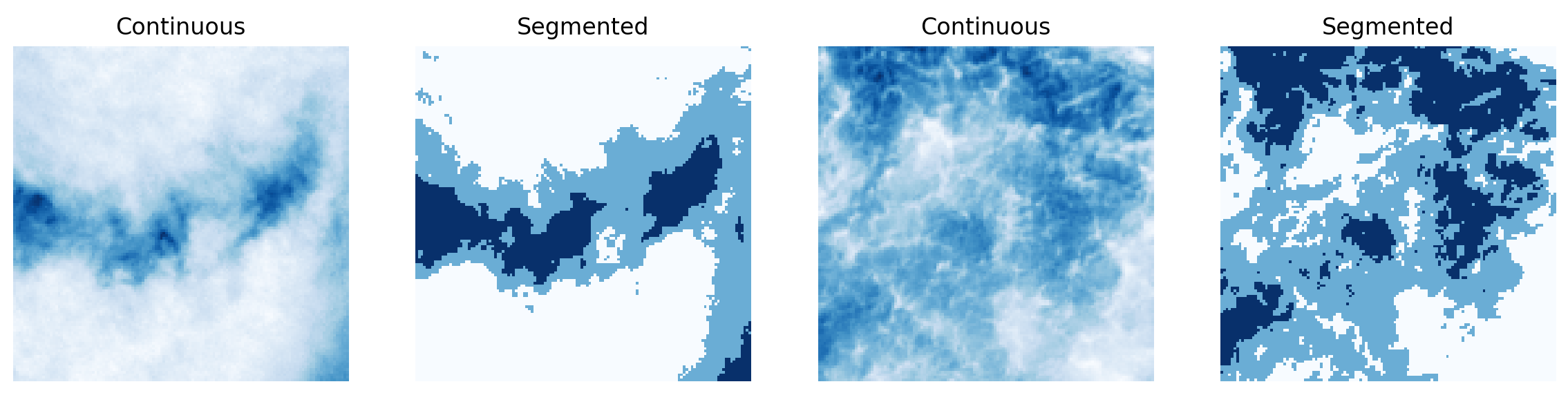}
    \caption{Cirrus band segmented by cirrus contamination level.}
\end{figure}
Visual observation supported the hypothesis that the K-Means clustering algorithm would find centroid values that account for the differing degrees of cirrus cloud contamination in the cirrus band. 
\subsection{Cirrus-to-Wildfire CNN}
\begin{figure}[H]
    \centering
    \includegraphics[scale=0.63]{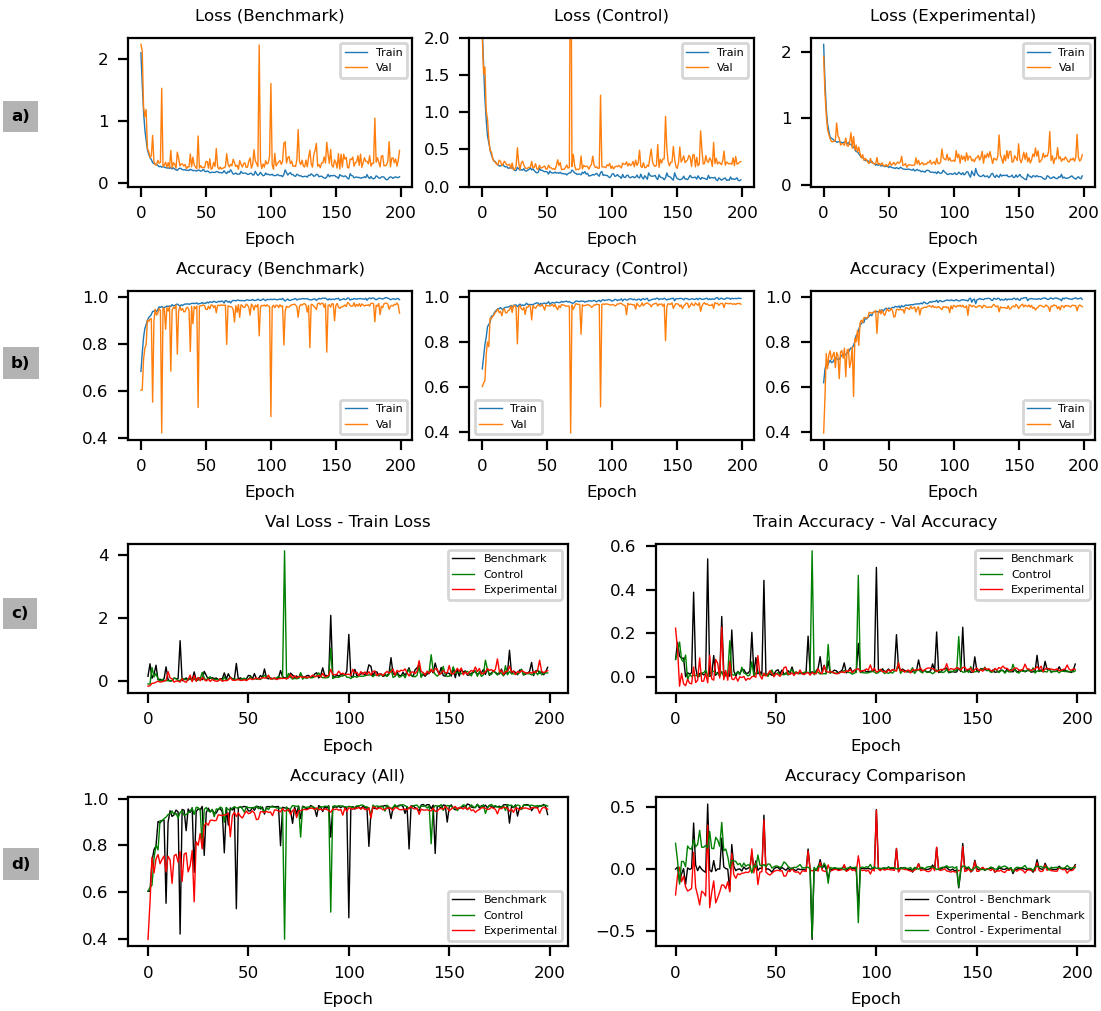}
    \caption{The learning curves of the three models used for the sensitivity analysis over 200 epochs.}
\end{figure}
\subsubsection{Learning Curves}
\paragraph{Loss}
Both the Benchmark model and the Control model experienced a high degree of noise during training. The Experimental model experienced much smoother training, but took longer to converge. 
\paragraph{Accuracy}
The Benchmark model and Control model reached 90\% validation accuracy around 25 epochs earlier than the Experimental model. However, the Experimental model maintained the plateau for around 165 more epochs while the validation accuracies of Benchmark and Control model fluctuated unpredictably. 
\paragraph{Overfitting}
All three models experienced minimal overfitting, but the difference between validation and train loss for the Experimental model trends upward at a faster rate compared to the Benchmark and Control models. This demonstrates a possible trade-off of the Experimental model. 
\paragraph{Comparison}
The Experimental model took longer to converge, but ultimately surpassed the accuracy of the Benchmark model while falling short to the accuracy of the Control model. The model metrics and train times are listed in the table below, where \textit{B} corresponds to Benchmark, \textit{C} corresponds to Control, and \textit{E} corresponds to Experimental.
\begin{table}[H]
\begin{center}
\renewcommand{\arraystretch}{1.3}
\caption{Performance Metrics for Sensitivity Analysis}
\begin{tabular}{|cclclcc|}
\hline
\multicolumn{1}{|c|}{\textbf{Hypotheses}}                                                                                            & \multicolumn{4}{c|}{\textbf{Metrics}}                                & \multicolumn{1}{c|}{\textbf{n}}                     & \textbf{\textit{P}-value}                    \\ \hline \hline
\multicolumn{7}{|c|}{\textbf{Binary Accuracy}}                                                                                                                                                                                                                               \\ \hline
\multicolumn{1}{|c|}{\multirow{2}{*}{\begin{tabular}[c]{@{}c@{}}$H_0: p_E - p_B = 0$\\ $H_A: p_E - p_B > 0$ \end{tabular}}}    &
\multicolumn{2}{c|}{$p_E$}      & \multicolumn{2}{c|}{$p_B$}      & \multicolumn{1}{c|}{\multirow{2}{*}{1,084}} & \multirow{2}{*}{0.00967}   \\ \cline{2-5}
\multicolumn{1}{|c|}{}                                                                                                   & \multicolumn{2}{c|}{0.95572} & \multicolumn{2}{c|}{0.93266} & \multicolumn{1}{c|}{}                      &                            \\ \hline

\multicolumn{1}{|c|}{\multirow{2}{*}{\begin{tabular}[c]{@{}c@{}}$H_0: p_E - p_C = 0$\\ $H_A: p_E - p_C < 0$\end{tabular}}}    & \multicolumn{2}{c|}{$p_E$}      & \multicolumn{2}{c|}{$p_C$}      & \multicolumn{1}{c|}{\multirow{2}{*}{1,084}} & \multirow{2}{*}{0.05750}   \\ \cline{2-5}
\multicolumn{1}{|c|}{}                                                                                                      & \multicolumn{2}{c|}{0.95572} & \multicolumn{2}{c|}{0.96863} & \multicolumn{1}{c|}{}                      &                            \\ \hline

\multicolumn{7}{|c|}{\textbf{Train Time (sec/epoch)}}                                                                                                                                                                                                                              \\ \hline
\multicolumn{1}{|c|}{\multirow{4}{*}{\begin{tabular}[c]{@{}c@{}}$H_0: \mu_E - \mu_B = 0$\\ $H_A: \mu_E - \mu_B > 0$\end{tabular}}} & \multicolumn{2}{c|}{$\mu_E$}      & \multicolumn{2}{c|}{$\mu_B$}      & \multicolumn{1}{c|}{\multirow{4}{*}{200}}  & \multirow{4}{*}{0.0}       \\ \cline{2-5}
\multicolumn{1}{|c|}{}                                                                                                      & \multicolumn{2}{c|}{2.76575} & \multicolumn{2}{c|}{2.57293} & \multicolumn{1}{c|}{}                      &                            \\ \cline{2-5}
\multicolumn{1}{|c|}{}                                                                                                      & \multicolumn{2}{c|}{$\sigma_E$}      & \multicolumn{2}{c|}{$\sigma_B$}      & \multicolumn{1}{c|}{}                      &                            \\ \cline{2-5}
\multicolumn{1}{|c|}{}                                                                                                      & \multicolumn{2}{c|}{0.05577} & \multicolumn{2}{c|}{0.06129} & \multicolumn{1}{c|}{}                      &                            \\ \hline
\multicolumn{1}{|c|}{\multirow{4}{*}{\begin{tabular}[c]{@{}c@{}}$H_0: \mu_E - \mu_C = 0$\\ $H_A: \mu_E - \mu_C < 0$\end{tabular}}}   & \multicolumn{2}{c|}{$\mu_E$}      & \multicolumn{2}{c|}{$\mu_C$}      & \multicolumn{1}{c|}{\multirow{4}{*}{200}}  & \multirow{4}{*}{4.6107E-8} \\ \cline{2-5}
\multicolumn{1}{|c|}{}                                                                                                      & \multicolumn{2}{c|}{2.76575} & \multicolumn{2}{c|}{2.77919} & \multicolumn{1}{c|}{}                      &                            \\ \cline{2-5}
\multicolumn{1}{|c|}{}                                                                                                      & \multicolumn{2}{c|}{$\sigma_E$}      & \multicolumn{2}{c|}{$\sigma_C$}      & \multicolumn{1}{c|}{}                      &                            \\ \cline{2-5}
\multicolumn{1}{|c|}{}                                                                                                      & \multicolumn{2}{c|}{0.05577} & \multicolumn{2}{c|}{0.06128} & \multicolumn{1}{c|}{}                      &                            \\\hline
\end{tabular}
\end{center}
\end{table}
\subsubsection{Sensitivity Analysis}
The superior binary accuracy of the Experimental model compared to the Benchmark was statistically significant at the $\alpha = 0.05$ level. However, the superior binary accuracy of the Control model compared to the Experimental model was not statistically significant. This shows that the binary accuracy of the Experimental model was not compromised by simplified cirrus features. \\\\\indent The faster train time of the Benchmark model compared to the Experimental was statistically significant, which was expected because the Benchmark model did not contain any cirrus inputs. The faster train time of the Experimental model compared to the Control model was also statistically significant. This shows that the Experimental model with simplified cirrus features provided a faster train time than the Control model, thereby decreasing computational expense.

\section{Discussion}
\subsection{Deep Learning Tasks}
\subsubsection{FCN} The results of the FCN show that despite the class imbalance inherent in the data set, the model is still a highly effective discriminator between fire and non-fire pixels. In fact, of all fire pixels in the validation data set, the model had a false negative rate of 1.46\%. This is a crucial trait for software used for real-time wildfire detection, since identifying regions with ground truth fires (i.e. maximizing recall) takes priority over minimizing false alarms. A potential use case for the FCN is Land Cover Change Detection. Aerial vehicles can periodically conduct forward propagation, accessing the predicted probability that each pixel of land contains a wildfire at that instant. Aerial vehicles can take snapshots of particular wildfire hot spots hourly, daily, etc. so that fluctuations in the pixel-wise probabilities could be analyzed using a time-series model to forecast possible spread.
\\\\
\indent Pereira et al. (2021) referred to their 34,525,121-parameter model as the ``U-Net'' and their 2,161,649-parameter model as the ``U-Net-Light'' \cite{pereira}. The table below compares the performance of these models to the performance of the FCN.
\begin{table}[H]
\caption{Performance Comparison}
\begin{center}
\footnotesize
\begin{tabular}{||c|c|c c c c||}
    \hline \\[-1em]
    Model & No. of Param. & P & R & $F_1$ & $F_2$ \\
    \hline\hline
    \\[-1em]
    ``U-Net'' & 34,525,121 & 0.898 & 0.888 & 0.893 & 0.890 \\ 
    \hline
    \\[-1em]
    ``U-Net-Light'' & 2,161,649 & 0.908 & 0.861 & 0.884 & 0.870 \\
    \hline
    \\[-1em]
    \textbf{FCN} & \textbf{2,138,785} & \textbf{0.878} & \textbf{0.985} & \textbf{0.928} & \textbf{0.962} \\ 
    \hline
\end{tabular}
\end{center}
\end{table}
The architecture of the FCN required less regularization than the ``U-Net'' and ``U-Net-Light,'' which used dropout layers between convolutions. This corroborates the findings of Sun et al. (2021), who demonstrated that high performance can be achieved using a simpler architecture with less or no regularization, compared to a complex architecture with heavy regularization \cite{dropout}.
\subsubsection{Cirrus-to-Wildfire CNN} The Experimental model had a lower computational expense than the Control model, yet did not significantly decrease the binary accuracy despite using simplified cirrus features. The success of the Experimental model shows that models with simpler features have the potential to detect active wildfires to similar degrees of accuracy than models with complex features, while also providing the added value of faster train times. It has already been established that not all Landsat 8 bands are needed to detect wildfires. The next step is to conduct feature engineering to find optimal simplifications of the input data. In this manner, less data will be needed to train a more versatile model that is better suited for South American wildfire hot spots with scarce computational resources.
\subsection{Relationship Between Cirrus Clouds and Wildfire Acreage} \label{cirrus eda}
Recall that the pixel values of the cirrus bands were segmented into three cirrus contamination categories: dense, scattered, and none. These segmented images were analyzed to find an association between the number of fire pixels and the number of pixels in each contamination category. Each scatter plot below contains 5,420 data points, with each point representing one image in the data set for the Cirrus-to-Wildfire CNN.
\begin{figure}[H]
    \centering
    \includegraphics[scale=0.3]{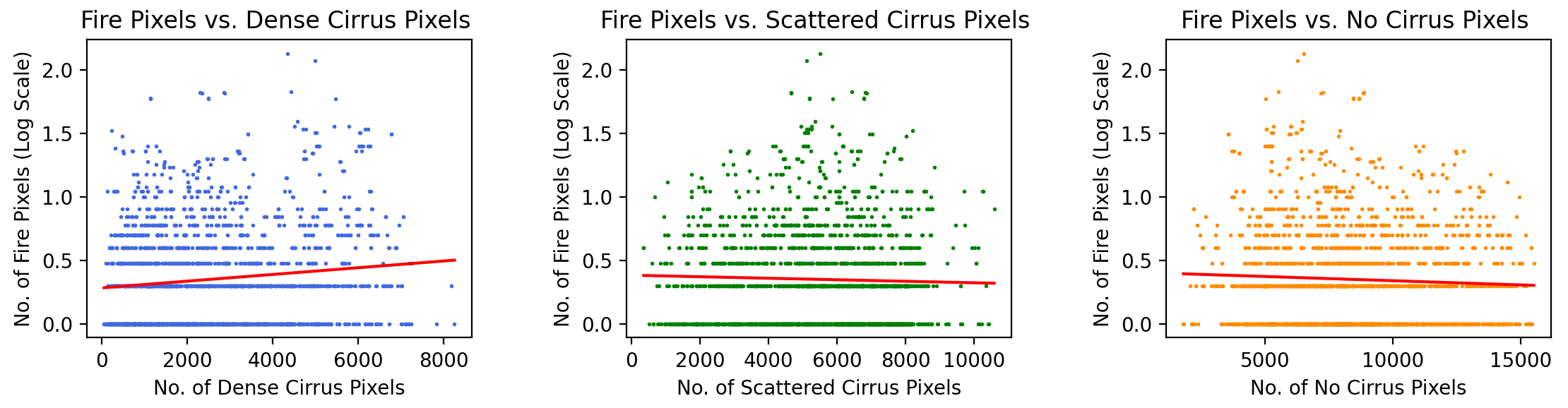}
    \caption{For each image, the number of fire pixels is plotted against the number of pixels in each cirrus contamination class. Linear regression fits are included to give a sense of form and direction.}
\end{figure}
From the figures above, it appears that the plot of Fire Pixels vs. Dense Cirrus Pixels is similar to a y-axis reflection of the plot of Fire Pixels vs. No Cirrus Pixels. Furthermore, it appears that the plot of Fire Pixels vs. Scattered Cirrus Pixels is nearly symmetric. Notice that the curves formed by the upper bound data points take on distinct shapes. The images with the largest number of fire pixels (points near the top of the graphs) appear on different regions from graph to graph, containing:
\begin{itemize}
    \itemsep-0.05em 
    \item More ``Dense Cirrus'' pixels than the mean number of ``Dense Cirrus'' pixels across all images.
    \item Similar numbers of “Scattered Cirrus” pixels as the mean number of ``Scattered Cirrus'' pixels across all images.
    \item Less ``No Cirrus'' pixels than the mean number of ``No Cirrus'' pixels across all images.
\end{itemize}
\indent These findings reveal the possibility of an underlying connection between a region's degree of cirrus contamination and probability of containing an active wildfire, supporting the hypothesis of Veselovskii et al. (2021)\cite{cirrus background}.
\section{Conclusions}
The computer vision models designed in this study are useful for wildfire classification and Land Cover Change Detection. A validation $F_2$ score of 0.962 and a high degree of generalizability, demonstrated by the test cases, prove the versatility of the fully convolutional model. The conclusions of the sensitivity analysis motivate avenues of future research, such as optimizing Landsat 8 feature engineering. Future research should also investigate the proposed relationship between the degree of cirrus cloud contamination and presence of a wildfire. 

\section*{Acknowledgement}
I would like to thank Mr. Michael Lordan for his time and feedback regarding this paper.


\begin{thebibliography}{1}
\bibitem{intro}
Aragão, L. E., Anderson, L. O., Fonseca, M. G., Rosan, T. M., Vedovato, L. B., Wagner, F. H., ... \& Saatchi, S. (2018). 21st Century drought-related fires counteract the decline of Amazon deforestation carbon emissions. Nature communications, 9(1), 1-12.

\bibitem{fn}
Arana-Pulido, V., Cabrera-Almeida, F., Perez-Mato, J., Dorta-Naranjo, B. P., Hernandez-Rodriguez, S., \& Jimenez-Yguacel, E. (2018).  Challenges of an Autonomous Wildfire Geolocation System Based on Synthetic Vision Technology. Sensors, 18(11), 3631.

\bibitem{batchnorm}
Dauphin, Y., \& Cubuk, E. D. (2020, September). Deconstructing the Regularization of BatchNorm. In International Conference on Learning Representations.

\bibitem{skip connections}
Drozdzal, M., Vorontsov, E., Chartrand, G., Kadoury, S., \& Pal, C. (2016). The importance of skip connections applications in biomedical image segmentation. In Deep learning and data labeling for medical applications (pp. 179-187). Springer, Cham.

\bibitem{khryashchev}
Khryashchev, V., \& Larionov, R. (2020, March). Wildfire Segmentation on Satellite Images using Deep Learning. In 2020 Moscow Workshop on Electronic and Networking Technologies (MWENT) (pp. 1-5). IEEE.

\bibitem{adam}
Kingma, D. P., \& Ba, J. (2014). Adam: A method for stochastic optimization. arXiv preprint arXiv:1412.6980.

\bibitem{landsat}
Knight, E. J., \& Knight, G. (2014). Landsat-8 operational land imager design, characterization and performance. Remote sensing, 6(11), 10286-10305.

\bibitem{matson holben}
Matson, M., \& Holben, B. (1987). Satellite detection of tropical burning in Brazil. International Journal of Remote Sensing, 8(3), 509-516.

\bibitem{normalization}
Patro, S., \& Sahu, K. K. (2015). Normalization: A preprocessing stage. arXiv preprint arXiv:1503.06462.

\bibitem{data set}
Pereira, G. H. D. A., Fusioka, A.M., Nassu, B. T., \& Minneto, R. (2020). A Large-Scale Dataset for Active Fire Detection/Segmentation (Landsat-8). IEEE Dataport. dx.doi.org/10.21227/t9gn-y009

\bibitem{pereira}
Pereira, G. H. D. A., Fusioka, A. M., Nassu, B. T., \& Minetto, R. (2021). Active Fire Detection in Landsat-8 Imagery: a Large-Scale Dataset and a Deep-Learning Study. arXiv preprint arXiv:2101.03409.

\bibitem{unet}
Ronneberger, O., Fischer, P., \& Brox, T. (2015, October). U-net: Convolutional networks for biomedical image segmentation. In International Conference on Medical image computing and computer-assisted intervention (pp. 234-241). Springer, Cham.

\bibitem{schroeder}
Schroeder, W., Oliva, P., Giglio, L., Quayle, B., Lorenz, E., \& Morelli, F. (2016). Active fire detection using Landsat-8/OLI data. Remote sensing of environment, 185, 210-220.

\bibitem{dropout}
Sun, C., Sharma, J., \& Maiti, M. (2021). Investigating the Relationship Between Dropout Regularization and Model Complexity in Neural Networks. arXiv preprint arXiv:2108.06628.

\bibitem{unsupervised semantic segmentation}
Van Gansbeke, W., Vandenhende, S., Georgoulis, S., \& Van Gool, L. (2021). Unsupervised semantic segmentation by contrasting object mask proposals. arXiv preprint arXiv:2102.06191.

\bibitem{cirrus background}
Veselovskii, I., Hu, Q., Ansmann, A., Goloub, P., Podvin, T., \& Korenskiy, M. (2021). Fluorescence lidar observations of wildfire smoke inside cirrus: A contribution to smoke-cirrus–interaction research. Atmospheric Chemistry and Physics Discussions, 1-29.
\end{thebibliography}
\end{document}